\documentclass[sigconf]{acmart}
\usepackage{amsmath}
\usepackage{graphicx}
\usepackage{placeins}

\AtBeginDocument{
  \providecommand\BibTeX{{
    \normalfont B\kern-0.5em{\scshape i\kern-0.25em b}\kern-0.8em\TeX}}}

\setcopyright{acmcopyright}
\copyrightyear{2022}
\acmYear{2022}
\acmDOI{00.0000/0000000.0000000}

\acmConference[NA '22]{NA '22}
\acmBooktitle{NA}

\begin{document}
\sloppy

\title{Visually Similar Products Retrieval for Shopsy}

\author{Prajit Nadkarni}
\email{prajit.pn@flipkart.com}
\affiliation{
    \institution{Flipkat Internet Pvt.\ Ltd.}
    \city{Bengaluru}
    \country{India}
}

\author{Narendra Varma Dasararaju}
\email{narendra.varma@flipkart.com}
\affiliation{
    \institution{Flipkat Internet Pvt.\ Ltd.}
    \city{Bengaluru}
    \country{India}
}


\begin{abstract}
    Visual search is of great assistance in reseller commerce, especially for non-tech savvy users with affinity towards regional languages.
    It allows resellers to accurately locate the products that they seek, unlike textual search which recommends products from head brands.
    Product attributes available in e-commerce have a great potential for building better visual search systems~\cite{Ak2021_FashionSearchNetV2, Serra2016, Liu_2016_DeepFashion} as they capture fine grained relations between data points.
    In this work, we design a visual search system for reseller commerce using a multi-task learning approach.
    We also highlight and address the challenges like image compression, cropping, scribbling on the image, etc, faced in reseller commerce.

    Our model consists of three different tasks: attribute classification, triplet ranking and variational autoencoder (VAE).
    Masking technique~\cite{Parekh2021} is used for designing the attribute classification.
    Next, we introduce an offline triplet mining technique which utilizes information from multiple attributes to capture relative order within the data.
    This technique displays a better performance compared to the traditional triplet mining~\cite{Schroff2015_Facenet} baseline, which uses single label/attribute information.
    We also compare and report incremental gain achieved by our unified multi-task model over each individual task separately.
    The effectiveness of our method is demonstrated using the in-house dataset of product images from the Lifestyle business-unit of Flipkart, India's largest e-commerce company.
    To efficiently retrieve the images in production, we use the Approximate Nearest Neighbor (ANN) index.
    Finally, we highlight our production environment constraints and present the design choices and experiments conducted to select a suitable ANN index.
\end{abstract}

\maketitle

\section{Introduction} \label{sec:introduction}
    Reseller commerce in India is a continuously growing multi-billion dollar market, which helps resellers utilize social platforms like Facebook and Whatsapp to bring commerce to the Next 500 Million customers.
    Resellers influence and assist their customers by curating the right products and doing order management, thus building a layer of trust and assistance for users to perform online shopping via social platforms.
    Shopsy is an app by Flipkart that allows resellers to share products with ease to their end customers and earn money by enabling commerce.
    Resellers communicate with the end user over social platforms in a way to keep all the Shopsy constructs hidden, thus building their business with no intermediary.
    Therefore, reseller communication with the end user does not include any product links, but is done entirely using images and written description.

    Reseller commerce covers following use cases:
    (1) Reseller promotes the products by sharing the images or description of the product with their end-user.
    The user then shows interest in buying a specific product and shares back the product image.
    (2) Reseller wants to check the availability of a product, that the user found interesting on social media, in the Shopsy catalog.
    It should be noted that images shared by the users with the reseller may be cropped or contain additional markings to highlight specific aspects of the product.
    Images also undergo compression while being shared over chat.
    The reseller may also add their logo on the image to promote their business.
    Over time as the reseller promotes multiple products to multiple users, it becomes hard to locate the requested product using textual search.
    It is difficult to describe visual characteristics of a product using words.
    Searching products using text tends to surface products from head brands and does not guarantee retrieval of the required item.
    Visual search tackles these limitations as it captures the exact visual patterns of the query image and retrieves the best matched item.
    Therefore, we build a visual search system for reseller commerce to handle the above mentioned use cases.

    In recent years, visual search has been built across many companies including Alibaba's Pailitao~\cite{Zhang2018}, Pinterest Flashlight and Lens~\cite{Zhai2017, Jing2015, Zhai2019_Pinterest}, Google Lens~\cite{Rajan2018}, Microsoft's Visual Search~\cite{Hu2018}, etc.
    These applications demonstrate large scale visual search systems for massive updating data.
    They focus on the quality of recommendations to improve user engagement.
    Flipkart catalog also contains millions of products and our primary aim is to assist the reseller in retrieving the exact item.
    The images in our catalog update upon the introduction of new products.
    Therefore, we design a system that offers high precision and low latency, while considering the size of our catalog and the update rate.

    In this work, we consider products only in the fashion category, as currently the reseller commerce in India is focused on fashion.
    Recent works in fashion~\cite{Ak2021_FashionSearchNetV2, Serra2016, Liu_2016_DeepFashion} have demonstrated the use of product attributes to build high quality visual embeddings using a combination of  attribute classification and triplet ranking loss.
    We design a multi-task model that learns from three different tasks: attribute-classification, triplet ranking and variational autoencoder.
    Finally, we highlight our production constraints and build an end to end visual search system for our use case.

    Our key contributions can be summarized as follows:
    \begin{itemize}
        \item We build a visual search system for reseller commerce and highlight challenges in this domain like image compression, cropping, scribbling on the image, etc.
        \item We present a triplet mining technique that uses information from multiple attributes to capture relative order within the data.
            It gives us twice as good performance as the traditional triplet mining technique, that uses a single label/attribute, which we have used as a baseline.
        \item We build a multi-task model to learn high-quality visual embeddings and attain a 4\% incremental gain over the best individual task.
        \item We highlight the business requirements and infrastructure constraints for our reseller commerce environment, and demonstrate an end to end visual search system that offers high precision and low latency, while considering our catalogue size and the data update rate.
        \item We present experiments and choices made for selecting an appropriate Approximate Nearest Neighbor (ANN) index for our production use case.
    \end{itemize}

\section{Related works}\label{sec:related-works}

    Large scale visual search systems have been built across many companies~\cite{Zhang2018, Zhai2017, Jing2015, Zhai2019_Pinterest, Rajan2018, Hu2018, Yang2017}, demonstrating large scale indexing for massive updating data.
    There has also been research in domain specific image retrieval systems, designed for fashion products~\cite{Ak2021_FashionSearchNetV2, Serra2016, Liu_2016_DeepFashion, Zhao2017, Innocente2021}.
    They leverage the product attribute information available in the e-commerce domain to build high quality visual embeddings.
    Other works that focus on extracting visual attributes for e-commerce~\cite{Parekh2021,Adhikari2019,Ferreira2018} demonstrate multi-class classification techniques.
    Parekh et al.\ \cite{Parekh2021} employ a masking technique to handle missing attribute values, a practical approach when dealing with products across different verticals.
    We use the same masking technique and build a multi-task learning approach with attribute classification and triplet ranking loss.

    Distance metric learning techniques are primarily designed for image retrieval systems, with the seminal works like contrastive-loss~\cite{Chopra2005_contrastive_loss} and triplet-loss~\cite{Schroff2015_Facenet}.
    Triplet loss considers a data point as anchor and associates it with a positive and a negative data point, and constrains the distance of an anchor-positive pair to be smaller than the anchor-negative pair.
    These methods have evolved over time, with early generations like Schroff et al.\ \cite{Schroff2015_Facenet}, where they introduced a semi-hard negative mining approach.
    This is an online triplet mining technique which computes useful triplets on the fly by sampling hard positive/negatives from within a mini-batch.
    Later, techniques evolved to incorporate information beyond a single triplet like Lifted Structured loss~\cite{Song2016}, N-Pair loss~\cite{Sohn2016}, etc.
    These losses associate an anchor point with a single positive and multiple negative points, and consider their relative hardness while pushing or pulling these points.
    The above losses consider the rich data-to-data relations and are able to learn fine-grained relations between them.
    However, these losses suffer from high training complexity $O(M^2)$ or $O(M^3)$ where M is the number of data points, thus slow convergence.
    Recent works like Proxy-NCA~\cite{Attias2017}, Proxy Anchor~\cite{Kim2020_ProxyAnchor}, etc, resolve the above complexity issue by introducing proxies, thus aiding in faster convergence.

    In all of the above losses, pair-based or proxy-based, the positives and negatives are chosen based on the class label, ie.\ positives are from the same class as anchor and negatives from a different class.
    For instance, in the face-recognition setting, to ensure enough positives in each mini-batch, Schroff et al.\ \cite{Schroff2015_Facenet} used a mini-batch of 1800 such that around 40 faces are selected per identity per mini-batch.
    In the case of proxy based losses, all proxies are part of the model and are kept in memory.
    Since each proxy represents a class, it puts a limit on the number of classes.
    Applying these techniques to e-commerce is challenging, where the possible class labels could be a product-id or a product-vertical (eg.\ t--shirt, shoe, watch, etc).
    In e-commerce, we have over millions of products with only 3--4 images per product, that appear on its product page, and the total number of verticals range only in a few hundreds.
    Choosing the class label as product-id can be too restrictive as there are only a few positives to learn from, and in the proxy based setting it would lead to millions of proxies.
    Choosing the product-vertical as class label makes the relation between data points too slack and thus we lose the fine grained intra-vertical details (e.g.\ discriminating one t--shirt pattern from another).
    Thus, applications in the e-commerce domain resort to using product attributes for mining the triplets.

    Ak et al.\ and others~\cite{Ak2021_FashionSearchNetV2, Innocente2021}, etc, choose triplets such that the anchor and the positive must have the same attribute value whereas the negative is chosen with a different attribute value.
    For instance, given that the anchor is a `blue' color, positive can be any image with `blue' color.
    Serra at el.\ ~\cite{Serra2016} use images with noisy tags (e.g.\ red-sweater, red-tshirt) and use similarity score `intersection over union' between the tags.
    They then choose positives that have a similarity score above a threshold and negatives with a score below the threshold.
    Shankar et al.\ ~\cite{Shankar2017} prepare triplets with three levels (positive, in-class-negative, out-of-class-negative), and use a `basic image similarity scorer' (e.g.\ pretrained AlexNet, color-histogram, PatternNet) for selecting candidates across levels.
    Drawing ideas from the above works, we define an offline triplet mining technique that prepares candidates across multiple levels, such that it captures the relative order within the data.
    We sample the candidates under each level based on the percentage of attributes matched.

    Another technique that has been used for image retrieval applications is Autoencoder~\cite{Hinton2006_autoencoder, Krizhevsky2011_AE2}.
    Autoencoder is a type of artificial neural network where the output is the same as the input.
    It has an encoder, a decoder and a bottleneck layer in the middle which captures the latent representation of the data.
    Thus, the bottleneck layer learns the most important characteristics of the image in an unsupervised way.
    A Variational Autoencoder (VAE)~\cite{Kingma2014} has the same structure as an autoencoder but uses a probabilistic approach to learn the latent representation.
    Unlike an autoencoder, VAE learns the disentangled embedding representation~\cite{Higgins2017_betaVAE}, i.e.\ where a single latent dimension is affected by only one generative factor and is invariant to changes in other factors.
    Thus, the underlying embedding spaces have a smooth continuous transformation over a latent dimension.
    For instance, a latent dimension which captures color variations, arranges the red t-shirt closer to maroon than a green t-shirt.
    This aspect can be beneficial in retrieving similar products along with the exact match.
    Sarmiento at el.\ ~\cite{Sarmiento2020} demonstrated the use of VAE for similar image retrieval of fashion products.

    We seek to combine the benefits of attribute-classification, triplet ranking and VAE to design a model for image retrieval.
    Ren at el.\ and others~\cite{Ren2017,Kendall2018} have shown performance gain of multi-task models over individual tasks.
    Kendall at el.\ and others~\cite{Kendall2018,Chen2018}  have explored ideas on balancing multiple loss objectives.
    In our work, we have taken the naive approach to combining multiple loss objectives and computed the linear sum of the normalized losses for each individual task.

\section{Dataset}\label{sec:dataset}

    \subsection{Datasets}\label{subsec:datasets}
\begin{figure}[!ht]
            \centering
            \includegraphics[width=\linewidth]{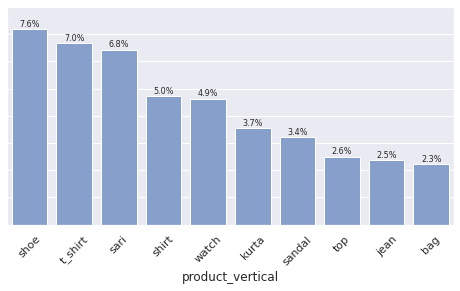}
            \caption{Distribution of top-10 product-verticals in the Lifestyle business-unit (1 year).
                Percentage denotes the proportion of vertical images in the dataset.
                The remaining verticals constitute 53.79\% of the dataset.
            }
            \label{fig:vertical_images}
        \end{figure}

        In this work, we use the in-house dataset of product-images from the Lifestyle business unit of Flipkart.
        To limit the size of the dataset for training, we consider only products that were ordered in the last one year.
        It has approximately 10+ million images, across 2+ million products spanning 250+ product verticals.
        A product has 3 to 4 images on average that are displayed on its product page.
        Figure-\ref{fig:vertical_images} shows the distribution of images for top product verticals.

        We create training, indexing, and querying subsets using the above dataset.
        These include:
        \begin{itemize}
            \item {\verb|lifestyle_1y|}: Contain images of all products from Lifestyle business-unit ordered in the last 1 year.
            \item {\verb|lifestyle_1y_train|} ($\sim1$ million images): Random 1M images sampled from {\verb|lifestyle_1y|} dataset.
                We split this into train-test-validation sets in the ratio 85:10:5.
            \item {\verb|lifestyle_1y_4v_index|} ($\sim3$ million images): We use this set to build an index for evaluating the image retrieval task, detailed in Section-\ref{subsec:image-retrieval-task}.
                The {\verb|lifestyle_1y|} dataset was too large to conduct multiple runs of experiments across different models.
                Therefore, we prepare this subset with all the images from top 4 verticals.
                We take `all the images' from these verticals (and not exclude images from {\verb|lifestyle_1y_train|})  to ensure that it represents the indexed data in the final production environment.
            \item {\verb|lifestyle_1y_4v_query|} (100k images): We use this as a test set to query the index, detailed in Section-\ref{subsec:image-retrieval-task}.
                This dataset contain random 100k images from {\verb|lifestyle_1y_4v_index|} dataset which are not present in training dataset {\verb|lifestyle_1y_train|}.
                The images in this dataset are augmented using augmentations described in Section-\ref{subsec:image-preprocessing}.
        \end{itemize}

    \subsection{Product Attributes}\label{subsec:product-attributes}
        For each image in the dataset, we assign a total of 12 attributes of its associated product.
        These are used for attribute classification task detailed in Section-\ref{subsubsec:attribute-classification-loss}.
        We have chosen 8 product-aspect attributes and 4 product-taxonomy attributes.
        There may be missing values for some of the product attributes, e.g.\ a \textit{shoe} will have no value for an attribute \textit{sleeve-length}.
        This is a common scenario when we deal with products across different verticals.
        Detailed list of attributes and their sample values are shown in the Table-\ref{tab:product_attributes}.
        \begin{table*}
            \caption{Product Attributes (for the training data of 1M images)}
            \label{tab:product_attributes}
            \begin{tabular}{llcl}
                \toprule
                Attribute Name          & Type         & Total Values  & Sample Values\\
                \midrule
                analytic\_category      & taxonomy     & 29            & WomenWesternCore, WomenEthnicCore, FashionJewellery, etc                     \\
                analytic\_sub\_category & taxonomy     & 77            & WesternWear, EthnicCore, WomenFashionJewellery, MensTShirt, Watch, etc       \\
                cms\_vertical           & taxonomy     & 147           & shoe, sari, t\_shirt, watch, shirt, etc                                      \\
                analytic\_vertical      & taxonomy     & 300           & WomenSari, MensRoundAndOthersTShirt, WomenKurtaAndKurti, etc                 \\
                color                   & aspect       & 33            & Black, Multicolor, Blue, White, Pink, etc                                    \\
                ideal\_for              & aspect       & 11            & Women, Men, Men \& Women, Girls, Boys, etc                                 \\
                material                & aspect       & 27            & Polyester, PU, Genuine Leather, Artificial Leather, Synthetic Leather, etc   \\
                occasion                & aspect       & 26            & Casual, Sports, Party \& Festive, Formal, Workwear, etc                    \\
                outer\_material         & aspect       & 23            & Synthetic, Mesh, Synthetic Leather, Leather, PU, etc                         \\
                pattern                 & aspect       & 39            & Solid, Printed, Self Design, Embroidered, Striped, etc                       \\
                sleeve                  & aspect       & 20            & Full Sleeve, Half Sleeve, Short Sleeve, 3/4 Sleeve, Sleeveless, etc          \\
                type                    & aspect       & 328           & Round Neck, Analog, Straight, Fashion, Polo Neck, etc                        \\
                \bottomrule
            \end{tabular}
        \end{table*}

    \subsection{Image Preprocessing}\label{subsec:image-preprocessing}
        We apply following augmentations to the images:
        \begin{itemize}
            \item cropping (random $180\times180$ crops on $224\times224$ image)
            \item color-augmentation (gray-scale, saturation, brightness)
            \item horizontal-flip
            \item rotation (0 to 90 degrees)
            \item overlay a $80\times80$ logo on top of $224\times224$ image
            \item jpeg-compression (quality: 20--50)
        \end{itemize}
        These are done to handle the image modifications introduced while sharing images over chat, with respect to the reseller use cases (Section-\ref{sec:introduction}).
        We also consider the possibilities of other augmentations that may happen in the photo editor like horizontal-flip, rotation, changes in brightness, saturation.

    \subsection{Triplet Generation}\label{subsec:image-triplets}
        We prepare triplets\emph{<anchor, positive, negative>} of images to learn the distance metric using triplet-loss, described in Section-\ref{subsubsec:triplet-loss}.
        Product-aspect attributes and product-vertical information are used to sample effective triplets to account for relative order in the data.
        We preprocess the dataset and prepare four candidate-levels for each anchor image, where each level contains a list of images:
        \begin{itemize}
            \item Level-0: list of images belonging to the \textit{same product} as that of the anchor image.
            \item Level-1: list of images belonging to products from the \textit{same vertical} as anchor image, and \textit{greater than 80\%} product-aspect attribute match with the product in anchor image.
            \item Level-2: list of images belonging to products from the \textit{same vertical} as anchor image, and \textit{less than 80\%} product-aspect attribute match with the product in anchor image.
            \item Level-3: list of images belonging to products from a \textit{different vertical} than that of anchor image.
        \end{itemize}
        To limit the complexity of the above process, random sampling is used and the size of the list in each level is limited to 10 (chosen based on training complexity and number of epochs).

        During training, first, an anchor-image is randomly sampled from the dataset to generate an image-triplet.
        A candidate positive image is then sampled by first randomly choosing a level out of 0,1,2 and then randomly choosing an image from the list of images in that level.
        To generate a negative image, we select the next consecutive level to that of the positively chosen level, and then randomly choose an image from the list of images in that level.
        For instance given an anchor image, we sample a positive from \textit{level-1} and the negative from the \textit{consecutive level-2}.
        We apply augmentations (Section-\ref{subsec:image-preprocessing}) to only the anchor image, and no augmentation is applied to the positive and negative images.

    \subsection{Training data}\label{subsec:training-data}
\begin{figure*}[!htbp]
            \centering
            \includegraphics[width=\textwidth]{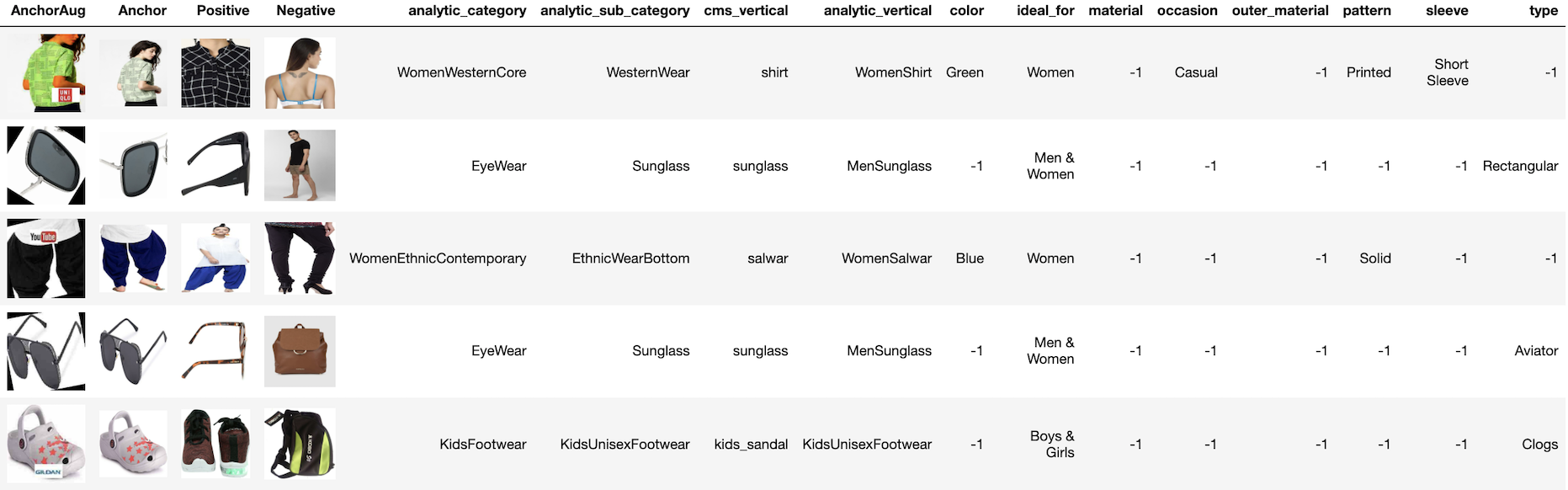}
            \caption{Random training samples.
                Only the anchor image is augmented which is shown in column AnchorAug.
                Note the similarity of positive and negative images with the anchor image, considering their levels mentioned in Section-\ref{subsec:image-triplets}.
                Columns on the right represent attributes of the product in anchor image, $-1$ denotes a missing attribute value for that product.
            }
            \label{fig:df_train}
        \end{figure*}

        For multi-task learning, we sample data by first choosing an image, and then club together corresponding product-attributes and image-triplets using the process described in above sections.
        Figure-\ref{fig:df_train} shows the combined data prepared for the multi-task model.

\section{Method}\label{sec:method}
    Our visual search system can be described in two parts:
    (1) First, a Convolutional Neural Network (CNN) is trained to generate embeddings that capture the notion of visual similarity.
    (2) Next, an Approximate Nearest Neighbor (ANN) Index is built to return similar images for the given query image embedding.
    We describe the embedding generation in Section-\ref{subsec:embedding-generation} and give the details of ANN Index in Section-\ref{subsec:ann-index}.

    \begin{figure}[!htbp]
        \centering
        \includegraphics[width=\linewidth]{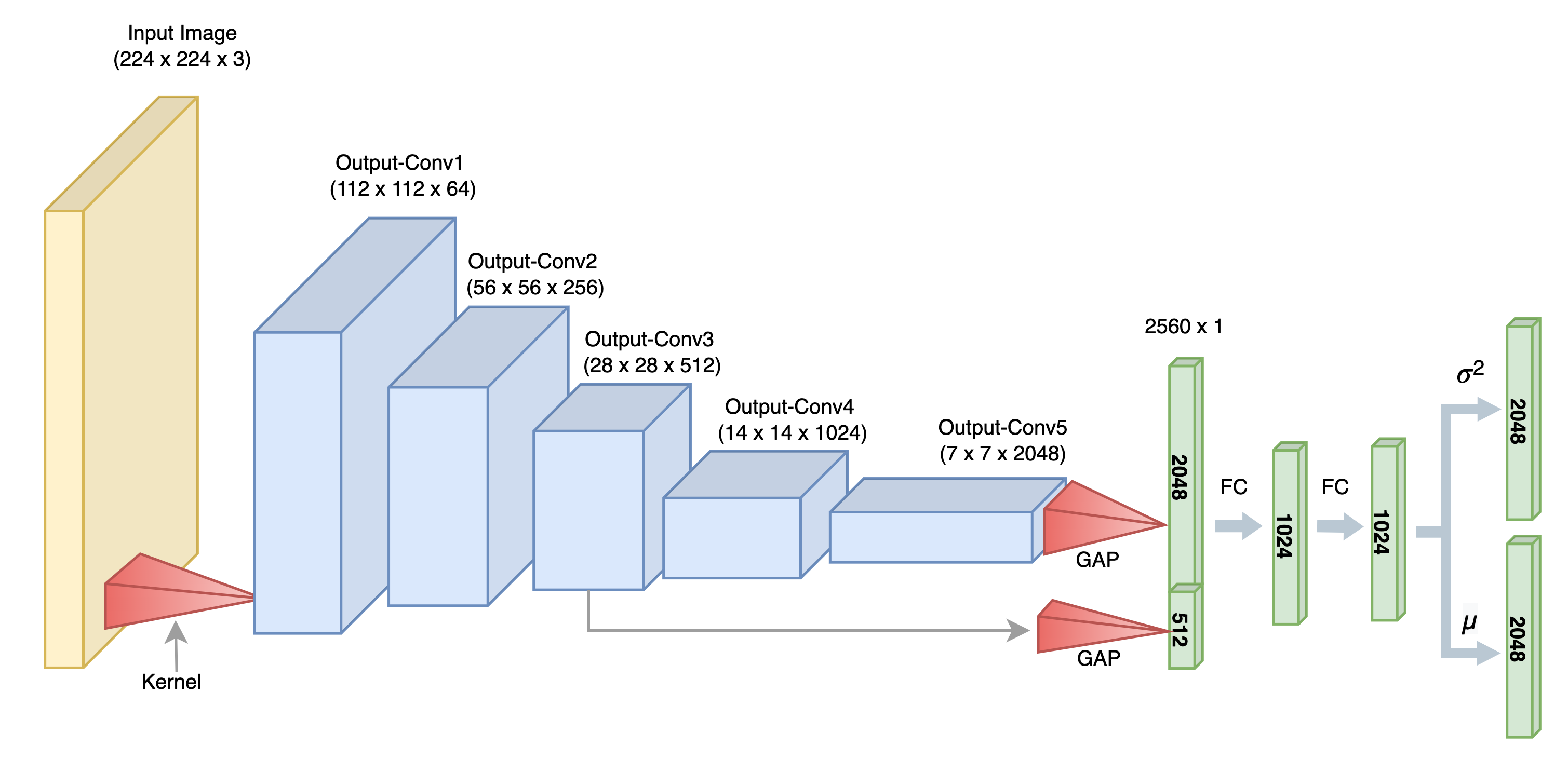}
        \caption{Encoder Network.
        We have used modified Resnet50~\cite{He2016}, where we concat the output of conv-block-3 with the final output from conv-block-5}
        \label{fig:encoder_model}
    \end{figure}
    \begin{figure*}[!htbp]
        \centering
        \includegraphics[width=\linewidth]{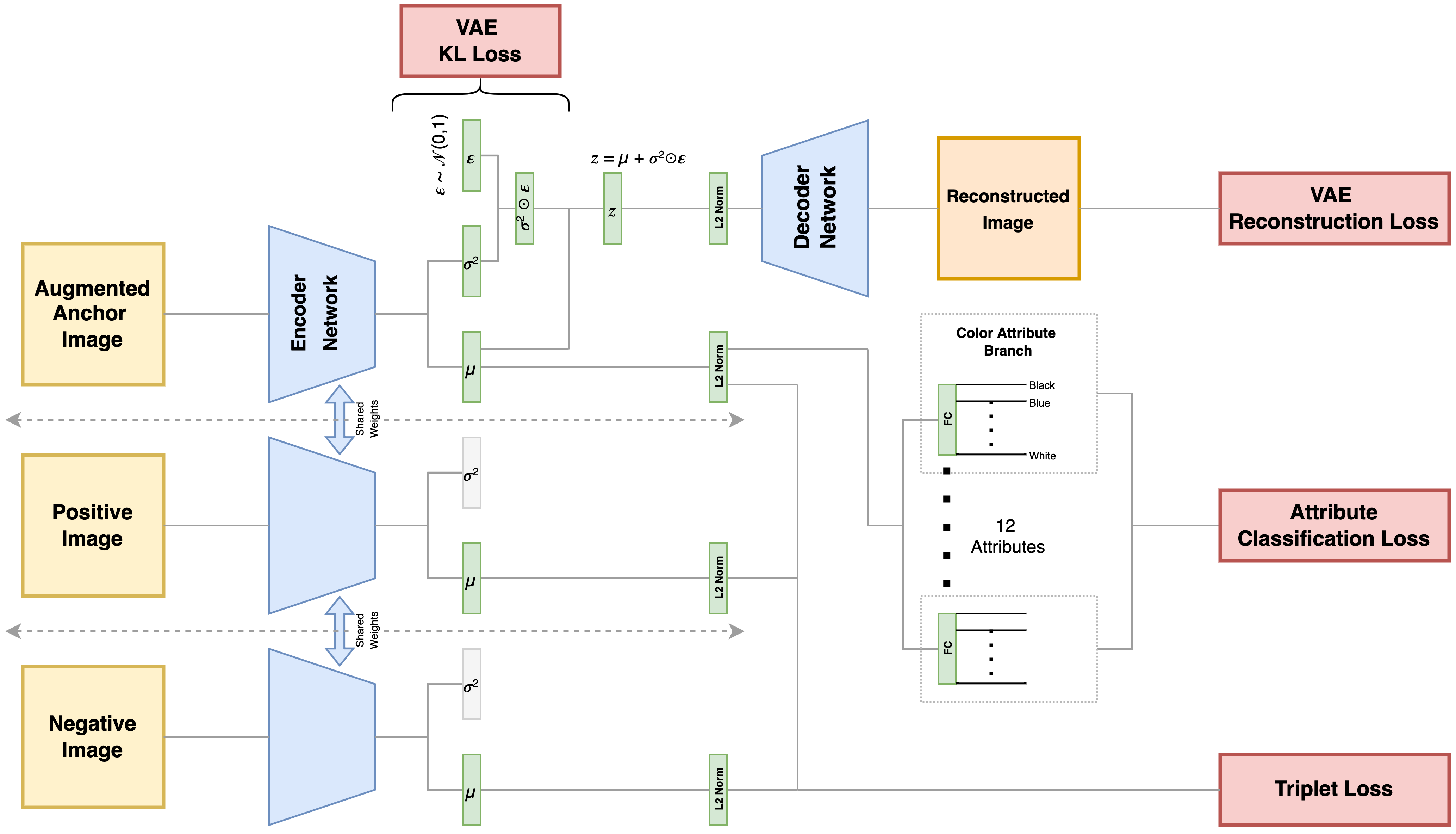}
        \caption{Multi-Task Model}
        \label{fig:multi_task_model}
    \end{figure*}

    \subsection{Embedding generation}\label{subsec:embedding-generation}
        \subsubsection{Backbone-CNN}\label{subsubsec:backbone-cnn}
            A modified version of Resnet50~\cite{He2016} is used as our backbone CNN model.
            We concat the output of conv-block-3 with the final output from conv-block-5 as shown in Figure-\ref{fig:encoder_model}.
            Our intuition is that, out of 5 blocks of convolution, the initial layers learn the fine-grained details (e.g.\ lines or blobs) while the later layers learn more abstract concepts like object class (e.g.\ shoe, t-shirt).
            Thus, the middle layers tend to capture the interesting aspects that may be able to distinguish within the same class of object (e.g.\ one t-shirt design from the other).

            As shown in Figure-\ref{fig:encoder_model}, the Resnet50 takes the input image of size $224\times224$.
            We use Global Average Pooling (GAP) to cast 2D conv maps to 1D\@.
            The pooled outputs from conv-block-5 and conv-block-3 are concatenated to get a 1D vector of length 2560.
            Further, 3 fully-connected layers (dim: 1024, 1024, 2048) are added with batch normalization and relu activation.

        \subsubsection{Attribute Classification Loss}\label{subsubsec:attribute-classification-loss}
            We use the embeddings generated by the above model to predict the product attributes.
            Thus, images with common product attributes will tend to have embedding vectors that are similar to each other.
            Our data contain multiple attributes that span across multiple product categories, thus not all attributes of a product are assigned values (e.g.\ sleeve-length is not applicable for shoes).
            To handle the missing values, we use the masking technique as mentioned in~\cite{Parekh2021}, where they mask the loss for the attributes with missing value.
            The model design for Attribute Classification task is shown in Figure-\ref{fig:multi_task_model}.

            For each attribute, we use multi-class prediction with a single label.
            12 product attributes are used as described in Section-\ref{subsec:product-attributes}.
            For each attribute branch, the embedding vector is passed to a dropout layer followed by a fully-connected layer with softmax activation.
            Then, categorical cross-entropy is used to compute loss for each attribute.
            Here, we mask the loss for an attribute with missing value.
            Finally, the loss across all attributes is averaged to compute the per sample loss for the attribute-classification task, which is given by:
            \begin{align*}
                Loss_{att\_clf}(y, \hat{y}) &= \frac{1}{A} \sum_{a=1}^{A} \ mask^a \times CCE(y^a, \hat{y}^a) \\
                mask^a &=
                \begin{cases}
                    0, & \text{if } y^a = -1\ (missing\ value\ for\ attribute\ a)\\
                    1, & \text{otherwise}
                \end{cases}
            \end{align*}
            where, A is the number of attributes, CCE is the categorical cross-entropy, $\hat{y}^a$ is the predicted label and $y^a$ is the ground truth label for attribute \emph{a}.
            For missing values, the ground truth label $y^a$ is set to $-1$.

        \subsubsection{Triplet Loss}\label{subsubsec:triplet-loss}
            In this task, we aim to learn the relative order in the data points by using a distance metric.
            We consider a triplet\emph{<anchor, positive, negative>} of images, which are sampled such that the anchor image is more similar to a positive image than a negative image (Section-\ref{subsec:image-triplets}).
            Then the embedding vector is learnt for each image, such that the distance between anchor-positive is less than the distance between anchor-negative, in the underlying embedding space.
            The embedding vectors are L2-normalized, which ensures that the embedding vectors are mapped to the surface of a n-dimensional hypersphere of radius 1.
            This enables us to use the euclidean distance as the similarity measure between the two vectors, since the euclidean distance over the surface of the unit hypersphere is bound between 0 to 2.
            We then compute the distance between anchor-positive and anchor-negative, which is then passed to the triplet-loss~\cite{Schroff2015_Facenet}.

            During the implementation, we take a triplet\emph{<anchor, positive, negative>} of images prepared in Section-\ref{subsec:image-triplets}, and pass each of the images through the encoder-network, as shown in the Figure-\ref{fig:multi_task_model}.
            The weights of the encoder-network are shared across the three paths.
            L2-normalization is then applied to the mean-$\mu$ layer output of the encoder-network.
            These L2-normalized image embeddings are used to calculate the squared euclidean distance.
            The model finally outputs two distances, i.e.\ distance between anchor-positive and anchor-negative, which are fed to a Hinge Loss function.
            The final loss from triplet task is given by:
            \[ Loss_{triplet}(a, p, n) = [d(a,p) - d(a,n) + M]_+ \]
            where a, p, n are the embedding vectors of anchor, positive and negative images respectively.
            The function $d(.)$ is the squared euclidean distance.
            M is the margin, and $[.]_+$ is the hinge function.
            We have used a margin of 0.2 (considering that squared euclidean distance ranges between 0 to 4 on the surface of n-dim hypersphere).

            We compare our triplet mining approach with triplet semi-hard mining technique~\cite{Schroff2015_Facenet} as the baseline.
            The semi-hard technique uses \textit{cms\_vertical} as the class label and samples the positives and negatives from within a mini-batch (size: 256), i.e.\ positives are chosen from the same class as the anchor and negatives from a different class.

        \subsubsection{VAE Loss}\label{subsubsec:vae-loss}
            In a variational autoencoder, there is an encoder, a decoder and a latent bottleneck layer in the middle.
            VAEs ensure latent embedding layer is normally distributed by using a new layer with parameters mean $\mu$ and standard deviation $\sigma$ of a normal distribution $N(\mu, \sigma^{2})$.
            During optimization, the normal distribution $N(\mu, \sigma^{2})$ is forced to be as close as possible to reference standard normal distribution $N(0,1)$ using Kullback-Leibler (KL divergence).
            In the process of reconstruction, the latent bottleneck layer learns the salient features of the image whilst the KL-divergence ensures the disentanglement~\cite{Higgins2017_betaVAE} within the learnt embedding space.
            Thus, independent latent units become sensitive to latent factors such as color, pattern, object-shape.
            And the embeddings learn a smooth continuous transformation over the values of these latent factors.
            For instance, all the t-shirts could lie in one dimension with gradual transition in their color.
            Our intuition is that this would cause the images in the embedding space to be arranged such that images with similar looking attributes lie near each other with gradual transition in the attribute value.
            Therefore, when the images are retrieved using nearest-neighbor algorithm, they are ordered with exact match followed by only a slight variation in the object characteristics.

            Our model design for variational autoencoder is shown in Figure-\ref{fig:multi_task_model}.
            Here, the encoder-network outputs the mean($\mu$) and log-variance($\log\sigma$) for a given input image.
            It is followed by a sampling layer which returns $z = \mu + \sigma^{2}\cdot\epsilon, \epsilon \sim N(0,1)$.
            The latent embedding is then L2-normalized, this is because downstream ANN indices work well if the vectors are L2-normalized.
            This is then passed to decoder-CNN\@.
            Our decoder-CNN architecture, shown in Appendix-\ref{sec:decoder-architecture}, uses transposed-convolutions with batch-normalization and relu activations.
            The last layer uses sigmoid activation to output the pixel values of the reconstructed image.
            We have used the approach from~\cite{Odena2016_deconvolution} to design the kernels for our decoder-CNN\@.
            This helps to reduce the checkerboard artifacts that appear in the reconstructed image.
            The end to end model takes an augmented image (Section-\ref{subsec:image-preprocessing}) as input, and it outputs the reconstructed image of exactly the same dimensions as the input image.

            The loss objective for VAE-Task is two fold.
            (1) For the reconstruction loss, we use binary cross-entropy~\cite{Kingma2014_SemiSup} between the original non-augmented input image and the reconstructed output image.
            (2) The KL-divergence loss~\cite{User3658307_2018} between the encoder distribution $q(z|x) = N(z|\mu(x), \Sigma(x))\ where\ \Sigma = diag(\sigma_1^2,\dots,\sigma_z^2)$ and the prior standard normal distribution $N(0,1)$.
            \begin{align*}
                Loss_{recon}(x, \hat{x}) &= \sum_{i=1}^{n} BCE(x_i, \hat{x}_i) \\
                Loss_{kl}(\mu, \sigma) &= \frac{1}{2} \bigg[ -\sum_{i=1}^{z} \big(\log\sigma_i^2 + 1\big) + \sum_{i=1}^{z}\sigma_i^2 + \sum_{i=1}^{z}\mu_i^2  \bigg]
            \end{align*}
            where BCE is binary cross-entropy, x, $\hat{x}$ are the input and the reconstructed images respectively, $n = 224\times224\times3$ is total image dimensions.
            $\mu, \sigma$ are the output of mean and variance layers respectively as shown in Figure-\ref{fig:multi_task_model}, $z = 2048$ is the number of latent dimensions.
            In practice, we output log-variance instead of the variance for numerical stability.

            We normalize each of the above loss values across their dimensions, and add them to get the final per sample loss for the VAE-Task.
            \[ Loss_{vae} = \frac{1}{n}Loss_{recon} + \frac{1}{z}Loss_{kl}  \]

        \subsubsection{Multi Task Loss}\label{subsubsec:multi-task-loss}
            Multi-Task learning aims to learn a single embedding representation of the given input image, which then outputs multiple values corresponding to different tasks.
            The overall setup for multi-task learning is shown in Figure-\ref{fig:multi_task_model}.
            The model predicts multiple outputs against each of the tasks:
            (1) For attribute classification task, it outputs softmax-probabilities of attribute-values against each of the 12 product-attributes using the augmented-anchor-image embedding.
            (2) For the triplet ranking task, it outputs two distances: distance between \emph{<augmented-anchor, positive>}, and the distance between \emph{<augmented-anchor, negative>}.
            (3) For VAE task, it takes the augmented-anchor-image as input, and outputs the reconstructed image with the same dimensions as the input image.
            We add the loss from each of the above tasks to construct the final per sample loss:
            \[ Loss_{multi\_task} = Loss_{att\_clf} + Loss_{triplet} + Loss_{vae} \]

        \subsubsection{Image Embedding}\label{subsubsec:image-embedding}
            To generate the embedding for a given query image using the above multi-task model, we take the output from mean-layer ($\mu$) and apply the L2-normalization.

    \subsection{ANN-Index}\label{subsec:ann-index}
        The only way to guarantee retrieval of exact nearest neighbors in n-dimensional space is exhaustive search, which is not practical to be used at query time in production.
        Thus, we resort to Approximate Nearest Neighbor (ANN) technique, which speeds up the search by preprocessing the data into efficient indices.
        There are many readily available libraries that implement ANN Indices e.g.\ Faiss~\cite{JDH2017_faiss}, ScaNN~\cite{AVQ2020_scann}, Annoy~\cite{Erik2018_annoy}, etc.
        Our primary aim is to assist the reseller in finding the exact item, thus we focus on high precision, followed by a throughput (queries-per-second) which ranges in few thousands.
        The total number of images in our system ranges in a few millions, and are updated only upon the introduction of new products in the catalog.
        Based on this business use case, we conducted exhaustive grid-search over multiple ANN indices, and found ScaNN~\cite{AVQ2020_scann} and HNSW~\cite{JDH2017_faiss} to work the best for our production use case.
        The detailed results of our experiments are presented in Section-\ref{subsec:selection-of-ann-index}.

        Given the embeddings generated in the previous section (Section-\ref{subsubsec:image-embedding}), we first use PCA to reduce the embedding dimensionality from 2048d to 256d.
        An ANN index is then built using the embeddings of all the images in our database.
        When a user searches using an image, the ANN index is queried using the image embedding and the top-k nearest neighbor images are retrieved.

    \subsection{Implementation}\label{subsec:implementation}
        We train our model using a batch-size of 32.
        Adam($\beta_1: .9, \beta_2: .999$) is used as an optimizer with a learning-rate of $10^{-3}$ for initial 100 epochs and then reduced to $10^{-5}$ till convergence.
        We end our training using early-stopping with patience of 10 epochs.
        Model is trained on single Tesla V100 GPU for a total of 148 epochs.

    \subsection{Evaluation}\label{subsec:evaluation}
        We evaluate our model on the image retrieval task, and use precision@k to measure the success.
        Precision@k returns the number of relevant results among the top k retrieved items.
        In production, we consider a query as successfully resolved if the exact item is present in the top k retrieved items, thus prec@k correctly measures our success.
        The detailed results are presented in Section-\ref{subsec:image-retrieval-task}.

\section{Experiments}\label{sec:experiments2}
    \begin{table}
        \caption{Retrieval Score (prec@4) per augmentation-type.
            For models (T-SH: triplet semi-hard, A: attribute-classification, T: triplet-loss, V: VAE, MT: multi-task)}
        \label{tab:image_retrieval_aug}
        \begin{tabular}{lrrrrr}
            \toprule
            Augmentation       & T-SH  & A & T & V & MT                      \\
            \midrule
            all\_augmentation  & 0.02        & 0.58               & 0.50                  & 0.10                  & \underline{0.64}              \\
            compression        & 0.83        & 0.97               & \underline{0.98}      & \underline{0.98}      & 0.97                          \\
            crop               & 0.12        & 0.85               & 0.85                  & 0.20                  & \underline{0.89}              \\
            hor\_flip          & 0.20        & 0.91               & \underline{0.95}      & 0.93                  & \underline{0.95}              \\
            no\_augmentation   & 0.98        & \underline{1.00}   & \underline{1.00}      & \underline{1.00}      & \underline{1.00}              \\
            logo\_overlay      & 0.62        & 0.97               & \underline{0.98}      & 0.95                  & \underline{0.98}              \\
            rotation           & 0.18        & 0.89               & 0.85                  & 0.77                  & \underline{0.93}              \\
            \midrule
            average            & 0.42        & 0.88               & 0.87                  & 0.70                  & \textbf{\underline{0.91}}     \\
            \bottomrule
        \end{tabular}
    \end{table}
    \begin{table*}
        \caption{ANN Index performance, ordered by precision then QPS(queries per second).
            The index uses a single cpu-core, and was built with total of 3 million images.
        }
        \label{tab:index_comparison}
        \begin{tabular}{lrrrrl}
            \toprule
            Index         & Precision@4         & QPS                   & Build Time (sec)  & Size(GB)  &  Hyper Params                                                    \\
            \midrule
            ScaNN         & \underline{0.87}    & \underline{999.19}    & 79	            & 3.37      & reorder\_num\_neighbors: 100, leaves\_to\_search: 50             \\
            HNSW\_Flat    & \underline{0.87}	& \underline{730.19}    & 313	            & 3.40      & efSearch: 100, efConstruction: 100, M: 16	                       \\
            HNSW\_SQ      & 0.87	            & 679.08                & 565	            & 1.16      & qtype: 0, efSearch: 100, efConstruction: 100, M: 16              \\
            IVF\_Flat     & 0.87	            & 102.28	            & 335	            & 3.01      & nprobe: 64, nlists: 10000                                        \\
            IVF\_SQ       & 0.87	            & 63.23	                & 383	            & 0.78      & qtype: 0, nprobe: 64, nlists: 10000                              \\
            Annoy         & 0.87	            & 62.25	                & 2354              & 6.43      & search\_k: 50000, prefault: True, n\_trees: 100                  \\
            FlatL2        & 0.87	            & 1.19                  & 4                 & 2.98      & -                                                                \\
            HNSW\_PQ      & 0.84                & 1,388.29              & 287               & 0.61      & pqM: 64, efSearch: 100, efConstruction: 100, M: 16               \\
            HNSW\_2Level  & 0.80                & 786.88                & 464               & 0.52      & pqM: 32, nlists: 1000, efSearch: 100, efConstruction: 100, M: 16 \\
            \bottomrule
        \end{tabular}
    \end{table*}

    \subsection{Image retrieval task}\label{subsec:image-retrieval-task}
        We set up the image-retrieval task such that it replicates the process in our production environment.
        In production, when a user searches using an image, the top-k items are retrieved from the ANN-index.
        We prepare two datasets, \textit{index-dataset} and \textit{query-dataset}.
        The {\verb|lifestyle_1y_4v_index|} (Section-\ref{subsec:datasets}) is used as \textit{index-dataset} which represents the images in our production database.
        The {\verb|lifestyle_1y_4v_query|} is used as the \textit{query dataset} which contain the augmented images and represent user queries that we may expect in our live system.

        First, we build the ANN index with the \textit{index-dataset} using the process described in Section-\ref{subsec:ann-index}.
        Next, for each image in the \textit{query-dataset}, top-k items are retrieved from the ANN-Index.
        If the query image-id is present in the top-k retrieved items, it is marked as success (1) else as a failure (0).
        Finally, we average over all the samples to compute precision@k.

\begin{figure}[h]
            \centering
            \includegraphics[width=\linewidth]{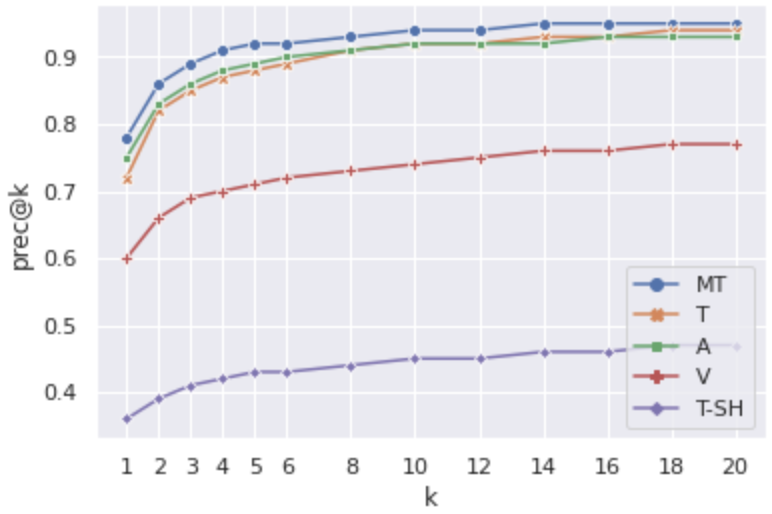}
            \caption{Image retrieval performance of models (T-SH: triplet semi-hard, A: attribute-classification, T: triplet-loss, V: VAE, MT: multi-task).
                We measure precision@k, where k is the top retrieved images from the ANN index.
            }
            \label{fig:image_retrieval_line}
            \Description{}
        \end{figure}

        We build separate models using each of the individual tasks (Attribute Classification, Triplet Loss, Triplet Semi-Hard, VAE) and compare it with our combined multi-task model and report the incremental gains achieved by multi-task technique.
        As shown in Figure-\ref{fig:image_retrieval_line}, the unified multi-task model performs better than the model learnt using each of the individual tasks.
        Also, we see that our triplet mining technique performs twice as better than the online triplet semi-hard mining~\cite{Schroff2015_Facenet} which uses \textit{cms\_vertical} as the class label.
        In Table-\ref{tab:image_retrieval_aug}, we compare the model performance for each individual image augmentation and display the results for prec@k=4.
        We see $4\% $ gain on average with the unified multi-task model over the best individual task performance.

    \subsection{Results of Individual Task}\label{subsec:results-of-individual-task}
        We also present the performance of each individual tasks within our multi-task model.
        Table-\ref{tab:att_clf_accuracy} displays the accuracy for the attribute prediction task against each individual attribute.
        Our final triplet-loss is $0.052$ where $77.9\%$ of samples are correctly ordered with zero triplet-loss.
        For the VAE task, the final reconstruction loss is $0.477$ and KL-divergence loss is $0.013$.
        Sample predictions for each task of the multi-task model are shown in Appendix-\ref{sec:multi-task-model-predictions}.
        \begin{table}
            \caption{Attribute Classification Accuracy}
            \label{tab:att_clf_accuracy}
            \begin{tabular}{lcl}
                \toprule
                Attribute Name          & Accuracy        \\
                \midrule
                analytic\_category      & 0.828           \\
                analytic\_sub\_category & 0.756           \\
                cms\_vertical           & 0.762           \\
                analytic\_vertical      & 0.646           \\
                color                   & 0.542           \\
                ideal\_for              & 0.852           \\
                material                & 0.543           \\
                occasion                & 0.703           \\
                outer\_material         & 0.407           \\
                pattern                 & 0.625           \\
                sleeve                  & 0.697           \\
                type                    & 0.555           \\
                \bottomrule
            \end{tabular}
        \end{table}

    \subsection{Selection of ANN Index}\label{subsec:selection-of-ann-index}
        Our goal is to choose the Approximate Nearest Neighbor (ANN) index that best suits our business use case, outlined in Section-\ref{subsec:ann-index}.
        Our production environment, detailed in Section-\ref{sec:production-environment}, demands an index that offers high precision, high QPS (queries-per-second), low memory footprint and preferably a short build time.
        Considering these requirements, we experimented with different indices (Table-\ref{tab:index_comparison}), and found ScaNN~\cite{AVQ2020_scann} and HNSW~\cite{JDH2017_faiss} to be well suited for our production use case.
        We also experimented with quantization techniques like IVF(Inverted File Index), PQ(Product Quantization) and SQ(Scalar Quantization), if supported in the index library.

        For this analysis, we build the index using {\verb|lifestyle_1y_4v_index|} dataset, and use {\verb|lifestyle_1y_4v_query|} as the query dataset.
        We chose to have only crop-augmented images since it represents the most common use case in our system.
        It also provides consistency in the image distribution while comparing multiple ANN Indices.
        We test around 3--4 values of each hyper-parameter using a grid search, and list the best configuration for each index in the Table-\ref{tab:index_comparison}.
        All indices are constrained to run on a single cpu-core.
        We can see from the Table-\ref{tab:index_comparison}, the top performing index(ScaNN, HNSW) experienced no drop in performance compared with exhaustive search (FlatL2).
        Further, the top index offers a QPS of $\sim1$k per cpu core and very modest memory footprint ($\sim3$GB / 3M images) which perfectly fits our production use case.

\section{Production Environment}\label{sec:production-environment}
    Considering the scale of reseller commerce in India and the available product discovery features (text search, content landing pages) in Shopsy, we estimate our visual search \textbf{QPS} to range in a few thousands during the peak business hours.
    Since our primary aim is to assist the reseller in locating the exact product, we focus on retrieving the images with \textbf{high precision}.
    Further, to ease the implementation process, we want to ensure our index fits on a single node (not distributed), thus limiting the \textbf{memory footprint} of the ANN index to be less than $\sim80$GB\@.
    The images in our system update upon introduction of new products in our catalog, which is not a frequent activity.
    But we prefer a short index \textbf{build time} because it adds convenience to the index updation process.
    Our top ANN index (Table-\ref{tab:index_comparison}) is able to satisfactorily address all the above constraints with minimal infrastructure demands.
    The embedding generation model (Section-\ref{subsubsec:image-embedding}) is able to process 9 QPS on a CPU core and 14 QPS on a GPU (since this is a single image and not batched).
    Thus, we aim to fulfill the business requirements by scaling the number of nodes (e.g.\ 5 CPU instances of 20 cores each, can support around a thousand QPS).
    Although we solve for a number of use cases like cropping, compression, etc, of catalog images, another important use case is `images in the wild', i.e.\ photos our users take with their cameras, which we would like to solve before we deploy the entire solution to production.

\section{Conclusion}\label{sec:conclusion}
    In this work, we build an end to end visual search system for reseller commerce, which helps to retrieve products more accurately compared to text search.
    We describe the evolution of distance metric learning over time and application of these methods to the e-commerce domain.
    We introduce an offline triplet mining technique which captures relative order within the data.
    Comparing our approach with the triplet semi-hard mining technique, which uses product-vertical as class label, shows twice as good performance as the latter.
    This indicates that attribute information plays a significant role when mining triplets for e-commerce.
    Further, we combine the benefits of multiple image retrieval approaches using multi-task learning and achieve a 4\% gain over the best individual learning approach.
    Finally, we highlight our business requirements and production environment constraints, and present the experiments conducted to select an ANN index that best suits our use case.

\FloatBarrier
\bibliographystyle{ACM-Reference-Format}
\bibliography{main}


\begin{thebibliography}{37}


\ifx \showCODEN    \undefined \def \showCODEN     #1{\unskip}     \fi
\ifx \showDOI      \undefined \def \showDOI       #1{#1}\fi
\ifx \showISBNx    \undefined \def \showISBNx     #1{\unskip}     \fi
\ifx \showISBNxiii \undefined \def \showISBNxiii  #1{\unskip}     \fi
\ifx \showISSN     \undefined \def \showISSN      #1{\unskip}     \fi
\ifx \showLCCN     \undefined \def \showLCCN      #1{\unskip}     \fi
\ifx \shownote     \undefined \def \shownote      #1{#1}          \fi
\ifx \showarticletitle \undefined \def \showarticletitle #1{#1}   \fi
\ifx \showURL      \undefined \def \showURL       {\relax}        \fi
\providecommand\bibfield[2]{#2}
\providecommand\bibinfo[2]{#2}
\providecommand\natexlab[1]{#1}
\providecommand\showeprint[2][]{arXiv:#2}

\bibitem[Adhikari et~al\mbox{.}(2019)]%
        {Adhikari2019}
\bibfield{author}{\bibinfo{person}{Sandeep~Singh Adhikari},
  \bibinfo{person}{Sukhneer Singh}, \bibinfo{person}{Anoop Rajagopal}, {and}
  \bibinfo{person}{Aruna Rajan}.} \bibinfo{year}{2019}\natexlab{}.
\newblock \bibinfo{title}{Progressive Fashion Attribute Extraction}.
\newblock
\newblock
\showeprint[arxiv]{1907.00157}~[cs.LG]


\bibitem[Ak et~al\mbox{.}(2021)]%
        {Ak2021_FashionSearchNetV2}
\bibfield{author}{\bibinfo{person}{Kenan~E. Ak}, \bibinfo{person}{Joo~Hwee
  Lim}, \bibinfo{person}{Ying Sun}, \bibinfo{person}{Jo~Yew Tham}, {and}
  \bibinfo{person}{Ashraf~A. Kassim}.} \bibinfo{year}{2021}\natexlab{}.
\newblock \bibinfo{title}{FashionSearchNet-v2: Learning Attribute
  Representations with Localization for Image Retrieval with Attribute
  Manipulation}.
\newblock
\newblock
\showeprint[arxiv]{2111.14145}~[cs.CV]


\bibitem[Bernhardsson(2018)]%
        {Erik2018_annoy}
\bibfield{author}{\bibinfo{person}{Erik Bernhardsson}.}
  \bibinfo{year}{2018}\natexlab{}.
\newblock \bibinfo{booktitle}{\emph{Annoy: Approximate Nearest Neighbors in
  C++/Python}}.
\newblock
\urldef\tempurl%
\url{https://pypi.org/project/annoy/}
\showURL{%
\tempurl}
\newblock
\shownote{Python package version 1.13.0}.


\bibitem[Chen et~al\mbox{.}(2018)]%
        {Chen2018}
\bibfield{author}{\bibinfo{person}{Zhao Chen}, \bibinfo{person}{Vijay
  Badrinarayanan}, \bibinfo{person}{Chen-Yu Lee}, {and} \bibinfo{person}{Andrew
  Rabinovich}.} \bibinfo{year}{2018}\natexlab{}.
\newblock \bibinfo{title}{GradNorm: Gradient Normalization for Adaptive Loss
  Balancing in Deep Multitask Networks}.
\newblock
\newblock
\showeprint[arxiv]{1711.02257}~[cs.CV]


\bibitem[Chopra et~al\mbox{.}(2005)]%
        {Chopra2005_contrastive_loss}
\bibfield{author}{\bibinfo{person}{S. Chopra}, \bibinfo{person}{R. Hadsell},
  {and} \bibinfo{person}{Y. LeCun}.} \bibinfo{year}{2005}\natexlab{}.
\newblock \showarticletitle{Learning a similarity metric discriminatively, with
  application to face verification}. In \bibinfo{booktitle}{\emph{2005 IEEE
  Computer Society Conference on Computer Vision and Pattern Recognition
  (CVPR'05)}}. \bibinfo{pages}{539--546 vol. 1}.
\newblock
\urldef\tempurl%
\url{https://doi.org/10.1109/CVPR.2005.202}
\showDOI{\tempurl}


\bibitem[D’Innocente et~al\mbox{.}(2021)]%
        {Innocente2021}
\bibfield{author}{\bibinfo{person}{Antonio D’Innocente},
  \bibinfo{person}{Nikhil Garg}, \bibinfo{person}{Yuan Zhang},
  \bibinfo{person}{Loris Bazzani}, {and} \bibinfo{person}{Michael Donoser}.}
  \bibinfo{year}{2021}\natexlab{}.
\newblock \showarticletitle{Localized Triplet Loss for Fine-grained Fashion
  Image Retrieval}. In \bibinfo{booktitle}{\emph{2021 IEEE/CVF Conference on
  Computer Vision and Pattern Recognition Workshops (CVPRW)}}.
  \bibinfo{pages}{3905--3910}.
\newblock
\urldef\tempurl%
\url{https://doi.org/10.1109/CVPRW53098.2021.00435}
\showDOI{\tempurl}


\bibitem[Ferreira et~al\mbox{.}(2018)]%
        {Ferreira2018}
\bibfield{author}{\bibinfo{person}{Beatriz~Quintino Ferreira},
  \bibinfo{person}{Luís Baía}, \bibinfo{person}{João Faria}, {and}
  \bibinfo{person}{Ricardo~Gamelas Sousa}.} \bibinfo{year}{2018}\natexlab{}.
\newblock \bibinfo{title}{A Unified Model with Structured Output for Fashion
  Images Classification}.
\newblock
\newblock
\showeprint[arxiv]{1806.09445}~[cs.CV]


\bibitem[Guo et~al\mbox{.}(2020)]%
        {AVQ2020_scann}
\bibfield{author}{\bibinfo{person}{Ruiqi Guo}, \bibinfo{person}{Philip Sun},
  \bibinfo{person}{Erik Lindgren}, \bibinfo{person}{Quan Geng},
  \bibinfo{person}{David Simcha}, \bibinfo{person}{Felix Chern}, {and}
  \bibinfo{person}{Sanjiv Kumar}.} \bibinfo{year}{2020}\natexlab{}.
\newblock \showarticletitle{Accelerating Large-Scale Inference with Anisotropic
  Vector Quantization}. In \bibinfo{booktitle}{\emph{International Conference
  on Machine Learning}}.
\newblock
\urldef\tempurl%
\url{https://arxiv.org/abs/1908.10396}
\showURL{%
\tempurl}


\bibitem[He et~al\mbox{.}(2016)]%
        {He2016}
\bibfield{author}{\bibinfo{person}{Kaiming He}, \bibinfo{person}{Xiangyu
  Zhang}, \bibinfo{person}{Shaoqing Ren}, {and} \bibinfo{person}{Jian Sun}.}
  \bibinfo{year}{2016}\natexlab{}.
\newblock \showarticletitle{Deep Residual Learning for Image Recognition}. In
  \bibinfo{booktitle}{\emph{2016 IEEE Conference on Computer Vision and Pattern
  Recognition (CVPR)}}. \bibinfo{pages}{770--778}.
\newblock
\urldef\tempurl%
\url{https://doi.org/10.1109/CVPR.2016.90}
\showDOI{\tempurl}


\bibitem[Higgins et~al\mbox{.}(2017)]%
        {Higgins2017_betaVAE}
\bibfield{author}{\bibinfo{person}{Irina Higgins}, \bibinfo{person}{Lo{\"i}c
  Matthey}, \bibinfo{person}{Arka Pal}, \bibinfo{person}{Christopher~P.
  Burgess}, \bibinfo{person}{Xavier Glorot}, \bibinfo{person}{Matthew~M.
  Botvinick}, \bibinfo{person}{Shakir Mohamed}, {and}
  \bibinfo{person}{Alexander Lerchner}.} \bibinfo{year}{2017}\natexlab{}.
\newblock \showarticletitle{beta-VAE: Learning Basic Visual Concepts with a
  Constrained Variational Framework}. In \bibinfo{booktitle}{\emph{ICLR}}.
\newblock


\bibitem[Hinton and Salakhutdinov(2006)]%
        {Hinton2006_autoencoder}
\bibfield{author}{\bibinfo{person}{Geoffrey~E Hinton} {and}
  \bibinfo{person}{Ruslan~R Salakhutdinov}.} \bibinfo{year}{2006}\natexlab{}.
\newblock \showarticletitle{Reducing the dimensionality of data with neural
  networks}.
\newblock \bibinfo{journal}{\emph{science}} \bibinfo{volume}{313},
  \bibinfo{number}{5786} (\bibinfo{year}{2006}), \bibinfo{pages}{504--507}.
\newblock


\bibitem[Hu et~al\mbox{.}(2018)]%
        {Hu2018}
\bibfield{author}{\bibinfo{person}{Houdong Hu}, \bibinfo{person}{Yan Wang},
  \bibinfo{person}{Linjun Yang}, \bibinfo{person}{Pavel Komlev},
  \bibinfo{person}{Li Huang}, \bibinfo{person}{Xi~(Stephen) Chen},
  \bibinfo{person}{Jiapei Huang}, \bibinfo{person}{Ye Wu},
  \bibinfo{person}{Meenaz Merchant}, {and} \bibinfo{person}{Arun Sacheti}.}
  \bibinfo{year}{2018}\natexlab{}.
\newblock \showarticletitle{Web-Scale Responsive Visual Search at Bing}. In
  \bibinfo{booktitle}{\emph{Proceedings of the 24th ACM SIGKDD International
  Conference on Knowledge Discovery \&; Data Mining}} (London, United Kingdom)
  \emph{(\bibinfo{series}{KDD '18})}. \bibinfo{publisher}{Association for
  Computing Machinery}, \bibinfo{address}{New York, NY, USA},
  \bibinfo{pages}{359–367}.
\newblock
\showISBNx{9781450355520}
\urldef\tempurl%
\url{https://doi.org/10.1145/3219819.3219843}
\showDOI{\tempurl}


\bibitem[Jing et~al\mbox{.}(2015)]%
        {Jing2015}
\bibfield{author}{\bibinfo{person}{Yushi Jing}, \bibinfo{person}{David Liu},
  \bibinfo{person}{Dmitry Kislyuk}, \bibinfo{person}{Andrew Zhai},
  \bibinfo{person}{Jiajing Xu}, \bibinfo{person}{Jeff Donahue}, {and}
  \bibinfo{person}{Sarah Tavel}.} \bibinfo{year}{2015}\natexlab{}.
\newblock \showarticletitle{Visual Search at Pinterest}. In
  \bibinfo{booktitle}{\emph{Proceedings of the 21th ACM SIGKDD International
  Conference on Knowledge Discovery and Data Mining}} (Sydney, NSW, Australia)
  \emph{(\bibinfo{series}{KDD '15})}. \bibinfo{publisher}{Association for
  Computing Machinery}, \bibinfo{address}{New York, NY, USA},
  \bibinfo{pages}{1889–1898}.
\newblock
\showISBNx{9781450336642}
\urldef\tempurl%
\url{https://doi.org/10.1145/2783258.2788621}
\showDOI{\tempurl}


\bibitem[Johnson et~al\mbox{.}(2017)]%
        {JDH2017_faiss}
\bibfield{author}{\bibinfo{person}{Jeff Johnson}, \bibinfo{person}{Matthijs
  Douze}, {and} \bibinfo{person}{Herv{\'e} J{\'e}gou}.}
  \bibinfo{year}{2017}\natexlab{}.
\newblock \showarticletitle{Billion-scale similarity search with GPUs}.
\newblock \bibinfo{journal}{\emph{arXiv preprint arXiv:1702.08734}}
  (\bibinfo{year}{2017}).
\newblock


\bibitem[Kendall et~al\mbox{.}(2018)]%
        {Kendall2018}
\bibfield{author}{\bibinfo{person}{Alex Kendall}, \bibinfo{person}{Yarin Gal},
  {and} \bibinfo{person}{Roberto Cipolla}.} \bibinfo{year}{2018}\natexlab{}.
\newblock \bibinfo{title}{Multi-Task Learning Using Uncertainty to Weigh Losses
  for Scene Geometry and Semantics}.
\newblock
\newblock
\showeprint[arxiv]{1705.07115}~[cs.CV]


\bibitem[Kim et~al\mbox{.}(2020)]%
        {Kim2020_ProxyAnchor}
\bibfield{author}{\bibinfo{person}{Sungyeon Kim}, \bibinfo{person}{Dongwon
  Kim}, \bibinfo{person}{Minsu Cho}, {and} \bibinfo{person}{Suha Kwak}.}
  \bibinfo{year}{2020}\natexlab{}.
\newblock \showarticletitle{Proxy Anchor Loss for Deep Metric Learning}. In
  \bibinfo{booktitle}{\emph{2020 IEEE/CVF Conference on Computer Vision and
  Pattern Recognition (CVPR)}}. \bibinfo{pages}{3235--3244}.
\newblock
\urldef\tempurl%
\url{https://doi.org/10.1109/CVPR42600.2020.00330}
\showDOI{\tempurl}


\bibitem[Kingma et~al\mbox{.}(2014)]%
        {Kingma2014_SemiSup}
\bibfield{author}{\bibinfo{person}{Diederik~P. Kingma},
  \bibinfo{person}{Danilo~J. Rezende}, \bibinfo{person}{Shakir Mohamed}, {and}
  \bibinfo{person}{Max Welling}.} \bibinfo{year}{2014}\natexlab{}.
\newblock \showarticletitle{Semi-Supervised Learning with Deep Generative
  Models}. In \bibinfo{booktitle}{\emph{Proceedings of the 27th International
  Conference on Neural Information Processing Systems - Volume 2}} (Montreal,
  Canada) \emph{(\bibinfo{series}{NIPS'14})}. \bibinfo{publisher}{MIT Press},
  \bibinfo{address}{Cambridge, MA, USA}, \bibinfo{pages}{3581–3589}.
\newblock


\bibitem[Kingma and Welling(2014)]%
        {Kingma2014}
\bibfield{author}{\bibinfo{person}{Diederik~P Kingma} {and}
  \bibinfo{person}{Max Welling}.} \bibinfo{year}{2014}\natexlab{}.
\newblock \bibinfo{title}{Auto-Encoding Variational Bayes}.
\newblock
\newblock
\showeprint[arxiv]{1312.6114}~[stat.ML]


\bibitem[Krizhevsky and Hinton(2011)]%
        {Krizhevsky2011_AE2}
\bibfield{author}{\bibinfo{person}{Alex Krizhevsky} {and}
  \bibinfo{person}{Geoffrey~E. Hinton}.} \bibinfo{year}{2011}\natexlab{}.
\newblock \showarticletitle{Using very deep autoencoders for content-based
  image retrieval}. In \bibinfo{booktitle}{\emph{ESANN}}.
\newblock


\bibitem[Liu et~al\mbox{.}(2016)]%
        {Liu_2016_DeepFashion}
\bibfield{author}{\bibinfo{person}{Ziwei Liu}, \bibinfo{person}{Ping Luo},
  \bibinfo{person}{Shi Qiu}, \bibinfo{person}{Xiaogang Wang}, {and}
  \bibinfo{person}{Xiaoou Tang}.} \bibinfo{year}{2016}\natexlab{}.
\newblock \showarticletitle{DeepFashion: Powering Robust Clothes Recognition
  and Retrieval With Rich Annotations}. In
  \bibinfo{booktitle}{\emph{Proceedings of the IEEE Conference on Computer
  Vision and Pattern Recognition (CVPR)}}.
\newblock


\bibitem[Movshovitz-Attias et~al\mbox{.}(2017)]%
        {Attias2017}
\bibfield{author}{\bibinfo{person}{Y. Movshovitz-Attias}, \bibinfo{person}{A.
  Toshev}, \bibinfo{person}{T.~K. Leung}, \bibinfo{person}{S. Ioffe}, {and}
  \bibinfo{person}{S. Singh}.} \bibinfo{year}{2017}\natexlab{}.
\newblock \showarticletitle{No Fuss Distance Metric Learning Using Proxies}. In
  \bibinfo{booktitle}{\emph{2017 IEEE International Conference on Computer
  Vision (ICCV)}}. \bibinfo{publisher}{IEEE Computer Society},
  \bibinfo{address}{Los Alamitos, CA, USA}, \bibinfo{pages}{360--368}.
\newblock
\showISSN{2380-7504}
\urldef\tempurl%
\url{https://doi.org/10.1109/ICCV.2017.47}
\showDOI{\tempurl}


\bibitem[Odena et~al\mbox{.}(2016)]%
        {Odena2016_deconvolution}
\bibfield{author}{\bibinfo{person}{Augustus Odena}, \bibinfo{person}{Vincent
  Dumoulin}, {and} \bibinfo{person}{Chris Olah}.}
  \bibinfo{year}{2016}\natexlab{}.
\newblock \showarticletitle{Deconvolution and Checkerboard Artifacts}.
\newblock \bibinfo{journal}{\emph{Distill}} (\bibinfo{year}{2016}).
\newblock
\urldef\tempurl%
\url{https://doi.org/10.23915/distill.00003}
\showDOI{\tempurl}


\bibitem[Parekh et~al\mbox{.}(2021)]%
        {Parekh2021}
\bibfield{author}{\bibinfo{person}{Viral Parekh}, \bibinfo{person}{Karimulla
  Shaik}, \bibinfo{person}{Soma Biswas}, {and} \bibinfo{person}{Muthusamy
  Chelliah}.} \bibinfo{year}{2021}\natexlab{}.
\newblock \showarticletitle{Fine-Grained Visual Attribute Extraction From
  Fashion Wear}. In \bibinfo{booktitle}{\emph{Proceedings of the IEEE/CVF
  Conference on Computer Vision and Pattern Recognition (CVPR) Workshops}}.
  \bibinfo{pages}{3973--3977}.
\newblock


\bibitem[Patel(2018)]%
        {Rajan2018}
\bibfield{author}{\bibinfo{person}{Rajan Patel}.}
  \bibinfo{year}{2018}\natexlab{}.
\newblock \bibinfo{title}{Google Lens: real-time answers to questions about the
  world around you}.
\newblock
\newblock
\urldef\tempurl%
\url{https://blog.google/products/google-lens/google-lens-real-time-answers-questions-about-world-around-you/}
\showURL{%
Retrieved Jan 11, 2022 from \tempurl}


\bibitem[Ren and Lee(2017)]%
        {Ren2017}
\bibfield{author}{\bibinfo{person}{Zhongzheng Ren} {and}
  \bibinfo{person}{Yong~Jae Lee}.} \bibinfo{year}{2017}\natexlab{}.
\newblock \bibinfo{title}{Cross-Domain Self-supervised Multi-task Feature
  Learning using Synthetic Imagery}.
\newblock
\newblock
\showeprint[arxiv]{1711.09082}~[cs.CV]


\bibitem[Sarmiento(2020)]%
        {Sarmiento2020}
\bibfield{author}{\bibinfo{person}{James-Andrew Sarmiento}.}
  \bibinfo{year}{2020}\natexlab{}.
\newblock \bibinfo{title}{Exploiting Latent Codes: Interactive Fashion Product
  Generation, Similar Image Retrieval, and Cross-Category Recommendation using
  Variational Autoencoders}.
\newblock
\newblock
\showeprint[arxiv]{2009.01053}~[cs.CV]


\bibitem[Schroff et~al\mbox{.}(2015)]%
        {Schroff2015_Facenet}
\bibfield{author}{\bibinfo{person}{Florian Schroff}, \bibinfo{person}{Dmitry
  Kalenichenko}, {and} \bibinfo{person}{James Philbin}.}
  \bibinfo{year}{2015}\natexlab{}.
\newblock \showarticletitle{FaceNet: A unified embedding for face recognition
  and clustering}. In \bibinfo{booktitle}{\emph{2015 IEEE Conference on
  Computer Vision and Pattern Recognition (CVPR)}}. \bibinfo{pages}{815--823}.
\newblock
\urldef\tempurl%
\url{https://doi.org/10.1109/CVPR.2015.7298682}
\showDOI{\tempurl}


\bibitem[Shankar et~al\mbox{.}(2017)]%
        {Shankar2017}
\bibfield{author}{\bibinfo{person}{Devashish Shankar}, \bibinfo{person}{Sujay
  Narumanchi}, \bibinfo{person}{H~A Ananya}, \bibinfo{person}{Pramod Kompalli},
  {and} \bibinfo{person}{Krishnendu Chaudhury}.}
  \bibinfo{year}{2017}\natexlab{}.
\newblock \bibinfo{title}{Deep Learning based Large Scale Visual Recommendation
  and Search for E-Commerce}.
\newblock
\newblock
\showeprint[arxiv]{1703.02344}~[cs.CV]


\bibitem[Simo-Serra and Ishikawa(2016)]%
        {Serra2016}
\bibfield{author}{\bibinfo{person}{Edgar Simo-Serra} {and}
  \bibinfo{person}{Hiroshi Ishikawa}.} \bibinfo{year}{2016}\natexlab{}.
\newblock \showarticletitle{Fashion Style in 128 Floats: Joint Ranking and
  Classification Using Weak Data for Feature Extraction}. In
  \bibinfo{booktitle}{\emph{2016 IEEE Conference on Computer Vision and Pattern
  Recognition (CVPR)}}. \bibinfo{pages}{298--307}.
\newblock
\urldef\tempurl%
\url{https://doi.org/10.1109/CVPR.2016.39}
\showDOI{\tempurl}


\bibitem[Sohn(2016)]%
        {Sohn2016}
\bibfield{author}{\bibinfo{person}{Kihyuk Sohn}.}
  \bibinfo{year}{2016}\natexlab{}.
\newblock \showarticletitle{Improved Deep Metric Learning with Multi-class
  N-pair Loss Objective}. In \bibinfo{booktitle}{\emph{Advances in Neural
  Information Processing Systems}}, \bibfield{editor}{\bibinfo{person}{D.~Lee},
  \bibinfo{person}{M.~Sugiyama}, \bibinfo{person}{U.~Luxburg},
  \bibinfo{person}{I.~Guyon}, {and} \bibinfo{person}{R.~Garnett}} (Eds.),
  Vol.~\bibinfo{volume}{29}. \bibinfo{publisher}{Curran Associates, Inc.}
\newblock
\urldef\tempurl%
\url{https://proceedings.neurips.cc/paper/2016/file/6b180037abbebea991d8b1232f8a8ca9-Paper.pdf}
\showURL{%
\tempurl}


\bibitem[Song et~al\mbox{.}(2016)]%
        {Song2016}
\bibfield{author}{\bibinfo{person}{Hyun~Oh Song}, \bibinfo{person}{Yu Xiang},
  \bibinfo{person}{Stefanie Jegelka}, {and} \bibinfo{person}{Silvio Savarese}.}
  \bibinfo{year}{2016}\natexlab{}.
\newblock \showarticletitle{Deep Metric Learning via Lifted Structured Feature
  Embedding}. In \bibinfo{booktitle}{\emph{Proceedings of the IEEE Conference
  on Computer Vision and Pattern Recognition (CVPR)}}.
\newblock


\bibitem[user3658307
  (https://stats.stackexchange.com/users/128284/user3658307)({[n.\,d.]})]%
        {User3658307_2018}
\bibfield{author}{\bibinfo{person}{user3658307
  (https://stats.stackexchange.com/users/128284/user3658307)}.}
  \bibinfo{year}{[n.\,d.]}\natexlab{}.
\newblock \bibinfo{title}{Deriving the KL divergence loss for VAEs}.
\newblock \bibinfo{howpublished}{Cross Validated}.
\newblock
\showeprint{https://stats.stackexchange.com/q/370048}
\urldef\tempurl%
\url{https://stats.stackexchange.com/q/370048}
\showURL{%
\tempurl}
\newblock
\shownote{URL:https://stats.stackexchange.com/q/370048 (version: 2020-03-03)}.


\bibitem[Yang et~al\mbox{.}(2017)]%
        {Yang2017}
\bibfield{author}{\bibinfo{person}{Fan Yang}, \bibinfo{person}{Ajinkya Kale},
  \bibinfo{person}{Yury Bubnov}, \bibinfo{person}{Leon Stein},
  \bibinfo{person}{Qiaosong Wang}, \bibinfo{person}{Hadi Kiapour}, {and}
  \bibinfo{person}{Robinson Piramuthu}.} \bibinfo{year}{2017}\natexlab{}.
\newblock \showarticletitle{Visual Search at EBay}. In
  \bibinfo{booktitle}{\emph{Proceedings of the 23rd ACM SIGKDD International
  Conference on Knowledge Discovery and Data Mining}} (Halifax, NS, Canada)
  \emph{(\bibinfo{series}{KDD '17})}. \bibinfo{publisher}{Association for
  Computing Machinery}, \bibinfo{address}{New York, NY, USA},
  \bibinfo{pages}{2101–2110}.
\newblock
\showISBNx{9781450348874}
\urldef\tempurl%
\url{https://doi.org/10.1145/3097983.3098162}
\showDOI{\tempurl}


\bibitem[Zhai et~al\mbox{.}(2017)]%
        {Zhai2017}
\bibfield{author}{\bibinfo{person}{Andrew Zhai}, \bibinfo{person}{Dmitry
  Kislyuk}, \bibinfo{person}{Yushi Jing}, \bibinfo{person}{Michael Feng},
  \bibinfo{person}{Eric Tzeng}, \bibinfo{person}{Jeff Donahue},
  \bibinfo{person}{Yue~Li Du}, {and} \bibinfo{person}{Trevor Darrell}.}
  \bibinfo{year}{2017}\natexlab{}.
\newblock \showarticletitle{Visual Discovery at Pinterest}. In
  \bibinfo{booktitle}{\emph{Proceedings of the 26th International Conference on
  World Wide Web Companion}} (Perth, Australia) \emph{(\bibinfo{series}{WWW '17
  Companion})}. \bibinfo{publisher}{International World Wide Web Conferences
  Steering Committee}, \bibinfo{address}{Republic and Canton of Geneva, CHE},
  \bibinfo{pages}{515–524}.
\newblock
\showISBNx{9781450349147}
\urldef\tempurl%
\url{https://doi.org/10.1145/3041021.3054201}
\showDOI{\tempurl}


\bibitem[Zhai et~al\mbox{.}(2019)]%
        {Zhai2019_Pinterest}
\bibfield{author}{\bibinfo{person}{Andrew Zhai}, \bibinfo{person}{Hao-Yu Wu},
  \bibinfo{person}{Eric Tzeng}, \bibinfo{person}{Dong Park}, {and}
  \bibinfo{person}{Charles Rosenberg}.} \bibinfo{year}{2019}\natexlab{}.
\newblock \showarticletitle{Learning a Unified Embedding for Visual Search at
  Pinterest}. \bibinfo{pages}{2412--2420}.
\newblock
\showISBNx{978-1-4503-6201-6}
\urldef\tempurl%
\url{https://doi.org/10.1145/3292500.3330739}
\showDOI{\tempurl}


\bibitem[Zhang et~al\mbox{.}(2018)]%
        {Zhang2018}
\bibfield{author}{\bibinfo{person}{Yanhao Zhang}, \bibinfo{person}{Pan Pan},
  \bibinfo{person}{Yun Zheng}, \bibinfo{person}{Kang Zhao},
  \bibinfo{person}{Yingya Zhang}, \bibinfo{person}{Xiaofeng Ren}, {and}
  \bibinfo{person}{Rong Jin}.} \bibinfo{year}{2018}\natexlab{}.
\newblock \showarticletitle{Visual Search at Alibaba}. In
  \bibinfo{booktitle}{\emph{Proceedings of the 24th ACM SIGKDD International
  Conference on Knowledge Discovery \&; Data Mining}} (London, United Kingdom)
  \emph{(\bibinfo{series}{KDD '18})}. \bibinfo{publisher}{Association for
  Computing Machinery}, \bibinfo{address}{New York, NY, USA},
  \bibinfo{pages}{993–1001}.
\newblock
\showISBNx{9781450355520}
\urldef\tempurl%
\url{https://doi.org/10.1145/3219819.3219820}
\showDOI{\tempurl}


\bibitem[Zhao et~al\mbox{.}(2017)]%
        {Zhao2017}
\bibfield{author}{\bibinfo{person}{Bo Zhao}, \bibinfo{person}{Jiashi Feng},
  \bibinfo{person}{Xiao Wu}, {and} \bibinfo{person}{Shuicheng Yan}.}
  \bibinfo{year}{2017}\natexlab{}.
\newblock \showarticletitle{Memory-Augmented Attribute Manipulation Networks
  for Interactive Fashion Search}. In \bibinfo{booktitle}{\emph{2017 IEEE
  Conference on Computer Vision and Pattern Recognition (CVPR)}}.
  \bibinfo{pages}{6156--6164}.
\newblock
\urldef\tempurl%
\url{https://doi.org/10.1109/CVPR.2017.652}
\showDOI{\tempurl}


\end{thebibliography}

\appendix

\begin{table*}
    \caption{Decoder architecture.
    Layer type represents TensorFlow layer types}
    \label{tab:decoder_architecture}
    \begin{tabular}{llllll}
        \toprule
        Layer Name          & Layer Type                & Kernel / stride     & Output Shape        & Num Params      & Connected to               \\
        \midrule
        input               & InputLayer                & -                   & (2048)              & 0               & -                             \\
        dense1              & Dense                     & -                   & (256)               & 524544          & input                         \\
        dense1\_bn          & BatchNormalization        & -                   & (256)               & 1024            & dense1                        \\
        dense1\_relu        & Activation                & -                   & (256)               & 0               & dense1\_bn                    \\
        dense2              & Dense                     & -                   & (50176)             & 12895232        & dense1\_relu                  \\
        dense2\_bn          & BatchNormalization        & -                   & (50176)             & 200704          & dense2                        \\
        dense2\_relu        & Activation                & -                   & (50176)             & 0               & dense2\_bn                    \\
        dense2\_reshape     & Reshape                   & -                   & (7, 7, 1024)        & 0               & dense2\_relu                  \\
        conv1               & Conv2DTranspose           & $2\times2\ /\ 2$    & (14, 14, 512)       & 2097664         & dense2\_reshape               \\
        conv1\_bn           & BatchNormalization        & -                   & (14, 14, 512)       & 2048            & conv1                         \\
        conv1\_relu         & Activation                & -                   & (14, 14, 512)       & 0               & conv1\_bn                     \\
        conv2               & Conv2DTranspose           & $2\times2\ /\ 2$    & (28, 28, 256)       & 524544          & conv1\_relu                   \\
        conv2\_bn           & BatchNormalization        & -                   & (28, 28, 256)       & 1024            & conv2                         \\
        conv2\_relu         & Activation                & -                   & (28, 28, 256)       & 0               & conv2\_bn                     \\
        conv3               & Conv2DTranspose           & $2\times2\ /\ 2$    & (56, 56, 128)       & 131200          & conv2\_relu                   \\
        conv3\_bn           & BatchNormalization        & -                   & (56, 56, 128)       & 512             & conv3                         \\
        conv3\_relu         & Activation                & -                   & (56, 56, 128)       & 0               & conv3\_bn                     \\
        conv3\_skip         & Conv2DTranspose           & $8\times8\ /\ 8$    & (56, 56, 128)       & 8388736         & dense2\_reshape               \\
        conv3\_add          & Add                       & -                   & (56, 56, 128)       & 0               & conv3\_relu, conv3\_skip      \\
        conv4               & Conv2DTranspose           & $2\times2\ /\ 2$    & (112, 112, 64)      & 32832           & conv3\_add                    \\
        conv4\_bn           & BatchNormalization        & -                   & (112, 112, 64)      & 256             & conv4                         \\
        conv4\_relu         & Activation                & -                   & (112, 112, 64)      & 0               & conv4\_bn                     \\
        conv5               & Conv2DTranspose           & $2\times2\ /\ 2$    & (224, 224, 3)       & 771             & conv4\_relu                   \\
        conv5\_bn           & BatchNormalization        & -                   & (224, 224, 3)       & 12              & conv5                         \\
        conv5\_sigmoid      & Activation                & -                   & (224, 224, 3)       & 0               & conv5\_bn                     \\
        \bottomrule
    \end{tabular}
\end{table*}
\begin{figure*}[!htpb]
    \centering
    \includegraphics[width=\linewidth]{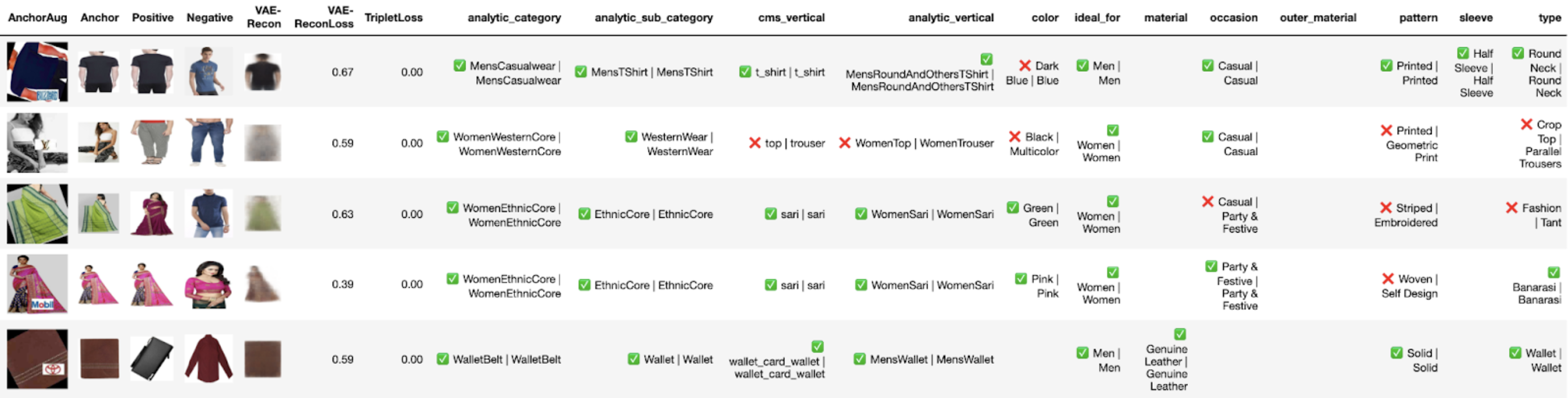}
    \caption{Sample predictions for each task of the multi-task model.
    Only the anchor image is augmented which is shown in column AnchorAug.
    It is followed by anchor, positive and negative images sampled using triplet mining technique (Section-\ref{subsec:image-triplets}).
    Column VAE-Recon shows the reconstructed anchor image (output of the decoder as shown in Figure-\ref{fig:multi_task_model}).
    Columns VAE-ReconLoss and TripletLoss show the loss values of the respective row.
    Columns on the right represent each of the product attributes, and the values are in the format \textit{<predicted-value>|<ground-truth-value>}.
    The green check highlights that attribute value is correctly predicted, while a red cross indicates that the prediction is incorrect.
    A blank cell indicates a missing attribute value for the product.
    }
    \label{fig:mt_predictions}
\end{figure*}

\section{Decoder architecture}\label{sec:decoder-architecture}
    Here we present our decoder architecture for the ease of reproducibility, shown in Table-\ref{tab:decoder_architecture}.

\section{Multi-Task model predictions}\label{sec:multi-task-model-predictions}
    Sample predictions for each task of the multi-task model are shown in Figure-\ref{fig:mt_predictions}.

\end{document}